\documentclass{article}
\usepackage{ijcai26}

\usepackage{times}
\usepackage{soul}
\usepackage{url}
\usepackage[hidelinks]{hyperref}
\usepackage[utf8]{inputenc}
\usepackage[small]{caption}
\usepackage{graphicx}
\usepackage{amsmath}
\usepackage{amsthm}
\usepackage{booktabs}
\usepackage{algorithm}
\usepackage{algorithmic}
\usepackage[switch]{lineno}
\usepackage{amssymb}
\usepackage{multirow} 
\usepackage{xcolor}

\title{SpatialV2A: Visual-Guided High-fidelity Spatial Audio Generation}

\author{
Yanan Wang$^1$
\and
Linjie Ren$^1$\and
Zihao Li$^1$\and
Junyi Wang$^1$\and
Tian Gan$^1$
\affiliations
$^1$School of Computer Science and Technology, Shandong University\\
}

\begin{document}

\maketitle

\begin{abstract}

While video-to-audio generation has achieved remarkable progress in semantic and temporal alignment, 
most existing studies focus solely on these aspects, 
paying limited attention to the spatial perception and immersive quality of the synthesized audio.
This limitation stems largely from current models' reliance on mono audio datasets, 
which lack the binaural spatial information needed to learn visual-to-spatial audio mappings.
To address this gap, we introduce two key contributions: 
we construct BinauralVGGSound, the first large-scale video-binaural audio dataset designed to support spatially aware video-to-audio generation; 
and we propose a end-to-end spatial audio generation framework guided by visual cues,
which explicitly models spatial features.
Our framework incorporates a visual-guided audio spatialization module that ensures the generated audio exhibits realistic spatial attributes and layered spatial depth while maintaining semantic and temporal alignment.
Experiments show that our approach substantially outperforms state-of-the-art models in spatial fidelity and delivers a more immersive auditory experience, without sacrificing temporal or semantic consistency.
The demo page can be accessed at https://github.com/renlinjie868-web/SpatialV2A.
\end{abstract}

\section{Introduction}
\label{intro}

Video-to-Audio (V2A) generation aims to synthesize high-fidelity, perceptually consistent audio signals that accurately match visual inputs. 
The core objective is to produce audio that is semantically aligned with visual events, temporally synchronized with scene dynamics, and perceptually realistic for an immersive user experience \cite{dagli2025see,jeong2025read,wang2024frieren,xing2024seeing}.
By establishing robust cross-modal associations between visual cues and their corresponding acoustic patterns, V2A systems can generate audio that accurately reflects object interactions, environmental context, and temporal motion cues.
Compared to manual Foley design, V2A offers higher production efficiency, lower labor costs, and scalability to large-scale video production~\cite{huang2025spotlighting}.  
As a result, V2A technology is increasingly important for applications such as video generation, virtual and augmented reality, and AI-assisted media production, where high-quality, visually grounded audio plays a crucial role in enhancing perceptual engagement and overall user experience.

\begin{table*}[t]
  \centering
  \caption{
  Comparison of existing audio-visual datasets. Many binaural datasets are small in scale, contain few categories, or lack captions, limiting their applicability to general V2A research. AV-C represents whether the audio signal and video scene are consistent.
  }
  \begin{tabular}{cccccccc}
  \hline
  \cline{3-8}   Dataset       &    Type   & Clips & Duration & Class & AV-C  &  Source & Caption \\
  \hline
  Fair-Play \cite{gao20192} & Binaural & 1.9k  & 5.2h & --     & \checkmark     & Indoor recording & -- \\
  SimBinaural \cite{garg2023visually} & Binaural & 22k   & 116.1h & 11    & \checkmark     & Simulation & -- \\
  YouTube-Binaural \cite{garg2023visually} & Binaural & 0.4k  & 27h   & --     & --     & Youtube & -- \\
  VGGSound \cite{chen2020vggsound} & Mono  & 199k  & 550h  & 309   & \checkmark     & Youtube & \checkmark \\
  \textbf{BinauralVGGSound (ours)} & Binaural & 187k  & 519h  & 309   & \checkmark     & Youtube & \checkmark \\
  \hline
  \end{tabular}%
  \label{tab_dataset}%
\end{table*}

\begin{figure}[t]
\centering
\includegraphics[width=0.45\textwidth]{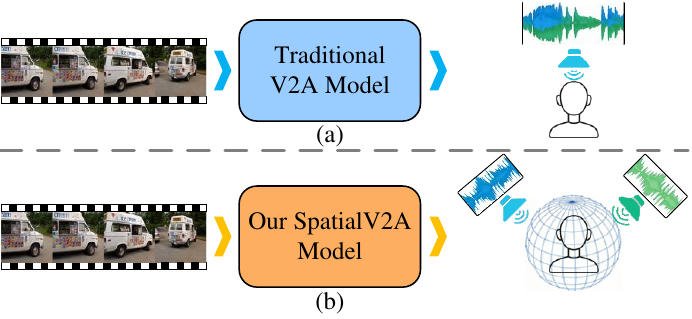}
\caption{
Comparison of traditional V2A models and the proposed SpatialV2A.
(a) Traditional V2A generates mono audio.
(b) SpatialV2A produces spatially consistent binaural audio.
}
\label{motivation_SpatialV2A}
\end{figure}

Although recent V2A studies have achieved remarkable progress in generating semantically and temporally aligned audio, they have largely overlooked audio spatialization \cite{jeong2025read,wang2024frieren,xing2024seeing}.
As illustrated in Figure \ref{motivation_SpatialV2A}(a), 
traditional V2A frameworks typically generate mono audio, which lacks directional and positional sound cues, thus failing to convey realistic spatial perception.
However, spatial audio plays a crucial role in human auditory perception by providing cues about sound source direction, position, and movement, which are essential for immersive and realistic audiovisual experiences \cite{pan2024innovative}.
Recent studies have begun to explore visual-guided audio spatialization, aiming to generate binaural or stereo audio from visual inputs \cite{dagli2025see}. 
Although these methods can synthesize spatialized audio, they often exhibit limited temporal and semantic alignment with visual events.
For example, See-2-Sound \cite{dagli2025see} generates spatial audio solely from visual content without leveraging the real audio signal, which may lead to inconsistencies between the synthesized sound and the visual scene.
Consequently, there is a critical need for a V2A framework that preserves semantic fidelity, temporal coherence, and spatial realism simultaneously, as illustrated by our proposed SpatialV2A in Figure \ref{motivation_SpatialV2A}(b).

Developing a spatially aware V2A framework faces two major challenges.
First, achieving spatial perception in V2A generation requires an explicit audio spatialization module that can deduce spatial sound cues directly from visual scenes.
Such a module must localize and temporally align sound-emitting events within the visual domain while capturing their relative intensity, thereby supplying the spatial information necessary for stereo-aware audio synthesis.
Second, the development of spatial V2A models is hindered by the limited availability of large-scale binaural datasets. 
Real-world binaural recordings requires specialized equipment and substantial human effort, making large-scale acquisition costly and impractical. 

As shwon in Table \ref{tab_dataset}, most existing visual–audio datasets are inadequate for training spatially aware V2A models: 
some provide only mono audio (\textit{e.g.}, VGGSound), 
while others are limited in scale, category diversity, or lack textual captions (\textit{e.g.}, Fair-Play and SimBinaural). 
These limitations underscore the need for a large-scale and diverse dataset that can leverage existing visual information to convert mono audio into a binaural format—thereby enabling the construction of extensive stereo visual–audio pairs to advance spatial V2A research.

To address these limitations, this work makes the following contributions:
\begin{enumerate}

\item We construct a large-scale spatial video–audio dataset, \textbf{BinauralVGGSound}, comprising over 187k high-quality video–audio pairs with binaural audio and descriptive captions. 
This dataset fills the gap of large-scale binaural-video datasets for V2A research, and establishes a data foundation for training and inference, and advancing spatially aware audio generation models.

\item We propose a spatially aware V2A generation framework \textbf{SpatialV2A} that models fine-grained spatial representations from visual inputs and synthesizes binaural stereo audio. 
Building upon the proposed dataset, our framework effectively integrates spatial perception into the V2A generation pipeline.

\item Extensive experiments demonstrate that our dataset construction strategy maintains the fidelity of the original mono audio, while the proposed framework generates binaural audio with precise temporal and semantic alignment. The results validate the effectiveness of both the dataset and the SpatialV2A framework for immersive audio-visual synthesis.
\end{enumerate}


\section{Related Work}
\label{sec:related_work}

\begin{figure*}
\centering
\includegraphics[width=0.95\textwidth]{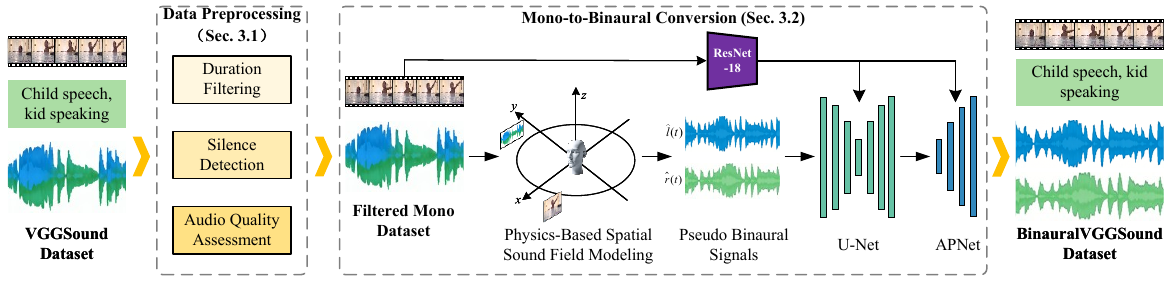}

\caption{
Overall pipeline of BinauralVGGSound dataset construction.
}
\label{data_pipeline}
\end{figure*}

\subsection{Video-to-Audio}

V2A generation aims to synthesize high-fidelity audio that is both semantically and temporally consistent with the visual content \cite{jeong2025read,luo2023diff}.
Leveraging the availability of pretrained multimodal models, 
recent studies have made substantial progress in improving both \textit{semantic} and \textit{temporal} alignment in V2A generation.

\textbf{Semantic Alignment.}
Semantic alignment ensure that the generated audio corresponds to the objects, actions, and scenes present in the video.
Most existing V2A approaches achieve this by utilizing large-scale pretrained multimodal encoders.
For instance, Im2Wav \cite{sheffer2023hear} and MMAudio \cite{jeong2025read} employ the CLIP encoder \cite{radford2021learning} to align visual and textual semantics, while Diff-Foley \cite{luo2023diff} introduces a contrastive audio-visual pretraining model to reinforce cross-modal correspondence.
These models provide a strong semantic foundation, which our method adopts to maintain consistent semantic alignment throughout the generation process.

\textbf{Temporal Alignment.}
Beyond semantic relevance, precise temporal alignment between visual events and audio signals is essential.
Recent works address this through various temporal feature synchronization mechanisms.
For example, SpecVQGAN \cite{iashin2021taming} conditions discrete audio representations on visual tokens to maintain temporal continuity. Diff-Foley \cite{luo2023diff} leverages contrastive audio-visual pretraining to learn temporally aligned cross-modal embeddings.
Furthermore, Synchformer \cite{iashin2024synchformer} improves fine-grained synchronization via attention-based cross-modal alignment.
Its effectiveness has been validated in subsequent works such as V-AURA \cite{viertola2025temporally}, MMAudio \cite{cheng2025mmaudio}, and ViSAudio \cite{zhang2025visaudio}, 
where integrating Synchformer significantly enhances audio–visual event alignment.

\subsection{Sound Source Localization} 
Sound source localization aims to identify the spatial coordinates of sounding regions in visual scenes by jointly analyzing audio signals and visual cues \cite{park2025hearing,sun2023learning,huang2023discovering}. 
Among state-of-the-art methods, 
ACL not only predicts precise sound source locations but also generates frame-level heatmaps that encode the relative intensity and spatial distribution of sounding regions \cite{park2025hearing}. 
These heatmaps provide rich spatial cues, including sound source location and intensity, which can be leveraged to extract spatial information for stereo audio generation.

\subsection{Audio Spatialization}

Recent advances in visual-guided audio spatialization have provided promising solutions for generating binaural audio \cite{xu2021visually}. 
Unlike traditional methods that rely solely on acoustic filters or Head-Related Transfer Functions \cite{zhang2011high}, 
visual-guided approaches \cite{CCStereo2025,xu2021visually,zhou2020sep} leverage visual information to infer scene geometry and the relative positions of sound sources.
This enables the transformation of monaural audio into binaural audio that better reflects the spatial layout of visual scenes.
A representative example is PseudoBinaural \cite{xu2021visually}, 
which synthesizes spatial audio by predicting interaural level and time differences from visual content, effectively transforming monaural audio into plausible stereo or binaural audio. 

\section{BinauralVGGSound Dataset Construction}
\label{sec:method}

\begin{table*}[htbp]
  \centering
  \caption{Analysis of the BinauralVGGSound dataset. Comparison of audio quality and spatialization with the VGGSound dataset.} 
  \renewcommand{\arraystretch}{1.2} 
    \begin{tabular}{c|ccc|ccccc}
    \hline
    \multirow{2}{*}{Dataset}  & \multicolumn{3}{c|}{Audio Quality} & \multicolumn{5}{c}{Audio Spatialization} \\
\cline{2-9}          & CE ↑    & CU ↑   & PQ  ↑  & IACC ↓ & ILD ↑  & ITD ↑   & ISD ↑  & IPD ↑\\
    \hline
    VGGSound \cite{chen2020vggsound} & 3.71  & 5.39    & 6.07  & -     & -     & -     & -     & - \\
    \textbf{BinauralVGGSound (ours)} & \textbf{3.74} & \textbf{5.44}  & \textbf{6.09} & \textbf{0.93}  & \textbf{3.08}  & \textbf{1.70}  & \textbf{0.07}  & \textbf{0.32}  \\
    \bottomrule
    \end{tabular}%
  \label{result_BinauralVGGSound}%
\end{table*}%

Constructing large-scale, visually aligned binaural audio is a critical prerequisite for training spatially aware V2A models. 
To this end, we introduce \textbf{BinauralVGGSound}, a large-scale binaural video-audio dataset for spatial V2A generation.
As illustrated in Figure \ref{data_pipeline}, the dataset construction pipeline consists of two main parts:
\textbf{Data Preprocessing} to ensure temporal alignment and audio quality; and 
\textbf{Mono-to-Binaural Conversion} to facilitate dual-channel spatial V2A modeling.

We adopted VGGSound \cite{chen2020vggsound} as the data source due to its broad category coverage, reliable audio-visual correspondence, and widespread adoption in prior V2A research \cite{jeong2025read,dagli2025see,xumanjie2024video,wang2024frieren,xing2024seeing}. 
The proposed pipeline bypasses the need for costly and hard-to-scale real-world binaural recordings by leveraging an existing mono video–audio dataset to synthesize binaural audio at scale. 
By doing so, BinauralVGGSound provides audio-visual aligned binaural data for spatial V2A modeling, dramatically lowering data acquisition costs and enabling scalable dataset construction.

\subsection{Data Preprocessing}
The audio-visual data preprocessing pipeline consists of three key steps:
(1) \textit{Duration filtering.} Clips shorter than 10~seconds are discarded to ensure sufficient temporal context for effective model training;
(2) \textit{Silence detection.} Videos whose audio contains over 80\% silence—identified via an audio activity detector—are removed to preserve meaningful audio–visual correspondence;
and (3) \textit{Audio quality assessment.} The Meta Audiobox Aesthetics Toolkit \cite{tjandra2025meta} is employed to filter out degraded or noisy audio samples, resulting in a high-quality dataset appropriate for spatial video-to-audio generation.
Detailed preprocessing statistics are provided in the supplementary material.

\subsection{Mono-to-Binaural Conversion}


Acquiring Ground-Truth (GT) binaural recordings paired with videos is prohibitively expensive and difficult to scale, severely limiting the size of existing datasets.
To overcome this, we convert the mono audio from VGGSound into spatially consistent binaural signals using the visual-guided audio spatialization framework PseudoBinaural \cite{xu2021visually}.

Specifically, for each mono audio signal $s(t)$, we first apply physics-based spatial sound field modeling to construct pseudo-binaural supervision. 
The mono signal is represented in the Spherical Harmonic (SH) domain and projected onto a set of virtual sound directions $\boldsymbol{\vartheta}_m'$, corresponding to the $m$-th spatial component, 
yielding spatial components $s_m^{\prime}(t) = \left(D^{\top} D\right)^{-1} D^{\top} \boldsymbol{\Psi}(t)$, 
where $D$ denotes the SH decoding matrix and $\boldsymbol{\Psi}(t)$ the spherical harmonic coefficients.

Pseudo-binaural signals are then obtained via Head-Related Impulse Response (HRIR)-based rendering:
\begin{equation}
\begin{aligned}
\hat{l}(t) &= \sum_{m=1}^M h_l\left(\boldsymbol{\vartheta}_m^{\prime}\right) \ast s_m^{\prime}(t)\text{,} \\
\hat{r}(t) &= \sum_{m=1}^M h_r\left(\boldsymbol{\vartheta}_m^{\prime}\right) \ast s_m^{\prime}(t)\text{,}
\end{aligned}
\end{equation}
where $M$ denotes the number of virtual loudspeakers, and $\ast$ is the convolution operation.

Based on the pseudo-binaural signals $(\hat{l}(t), \hat{r}(t))$, a visual-guided binaural generation model is trained following the PseudoBinaural paradigm. 
The model employs a U-Net backbone network \cite{ronneberger2015u} operating in the short-time Fourier transform domain, 
injects ResNet-18 \cite{he2016deep} visual features into the bottleneck layers,
and incorporates APNet-based separation modules \cite{zhou2020sep} to improve sound source discrimination. 
Once trained, the model is applied to the entire mono VGGSound dataset to generate spatially consistent binaural audio $(x^{l}, x^{r})$. 
The resulting model-generated binaural audio, together with the original videos and captions, 
constitutes the final BinauralVGGSound dataset.
Its scale and diversity, including the number of clips, total duration, and class coverage, are summarized in Table \ref{tab_dataset}.

\subsection{Data Validation}
To validate the effectiveness of the constructed BinauralVGGSound dataset, 
we conduct a comprehensive consistency analysis focusing on \textit{audio quality} and \textit{spatial stereo} characteristics.
For \textit{audio quality}, we adopt the Meta Audiobox Aesthetics Toolkit, which assesses three dimensions: Production Quality (PQ), Content Enjoyment (CE), and Content Usefulness (CU).
%
For \textit{spatial stereo}, we compute five widely used binaural metrics: 
Interaural Cross-Correlation (IACC), 
Interaural Level Difference (ILD), 
Interaural Time Difference (ITD), 
Interaural Spectral Difference (ISD), 
and Interaural Phase Difference (IPD) \cite{hernandez2024interaural}.

\begin{figure}[h]
\centering
\includegraphics[width=0.4\textwidth]{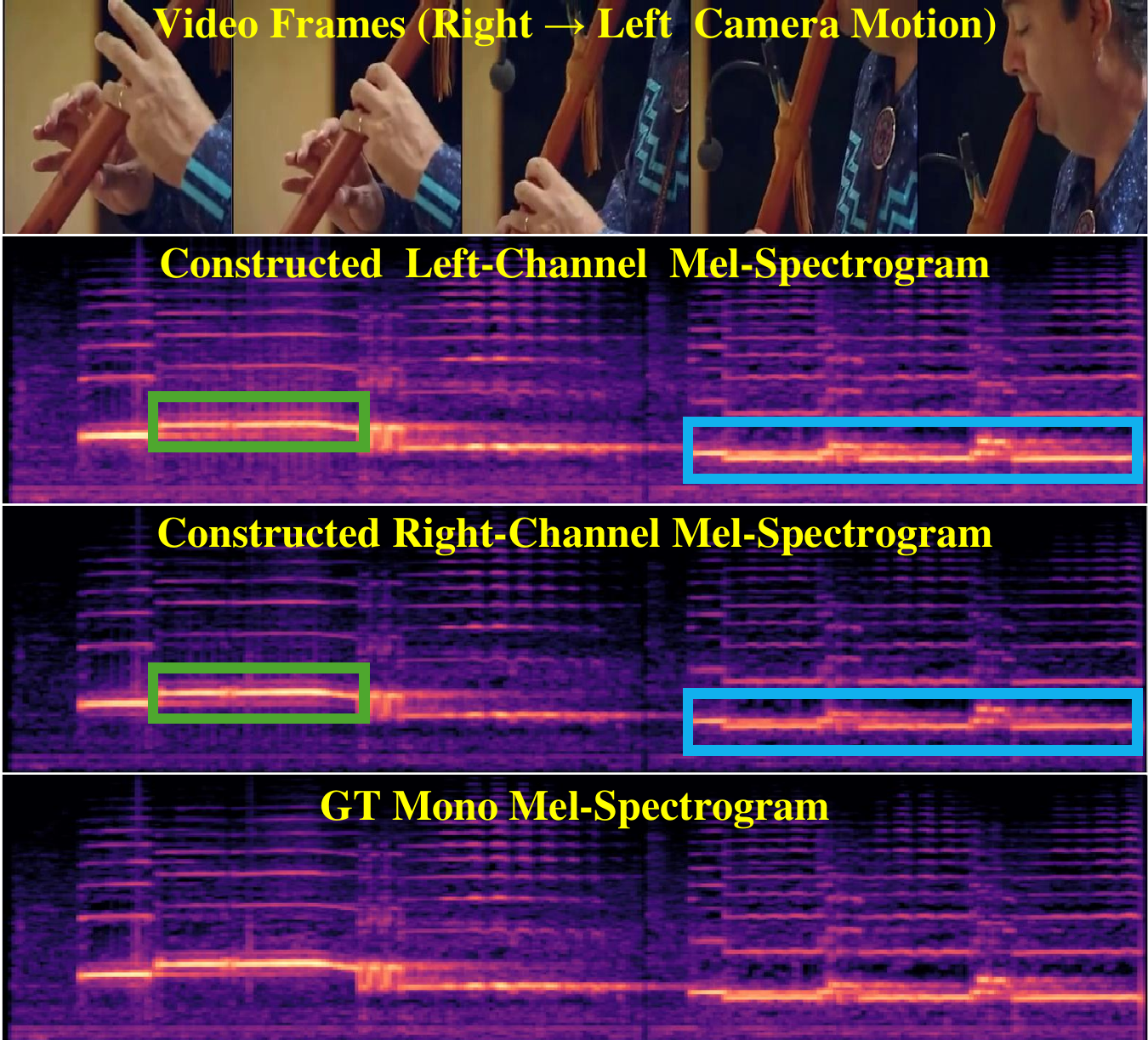}
\caption{
Example visualization from BinauralVGGSound for a flute-playing scene.
The transition of bounding boxes from green to blue reflects the estimated sound source movement, demonstrating spatial consistency with the visual motion.
}
\label{data_validation}
\end{figure}

As shown in Table \ref{result_BinauralVGGSound}, 
BinauralVGGSound maintains audio quality scores comparable to—and in some cases slightly higher than—the original VGGSound. 
More importantly, it introduces well-defined spatial cues across all binaural metrics, which are entirely absent in the mono original.
%
%
Furthermore, Figure \ref{data_validation} illustrates that spatial variations in the binaural audio consistently correspond with visual motion, confirming accurate audio–visual spatial alignment.
Additional validation examples are provided in the supplementary material.
%

These results demonstrate that our construction pipeline successfully enriches the audio with realistic spatial attributes while preserving high perceptual quality, thereby validating the dataset and establishing a solid foundation for spatially aware V2A research.

\begin{figure*}
\centering
\includegraphics[width=\textwidth]{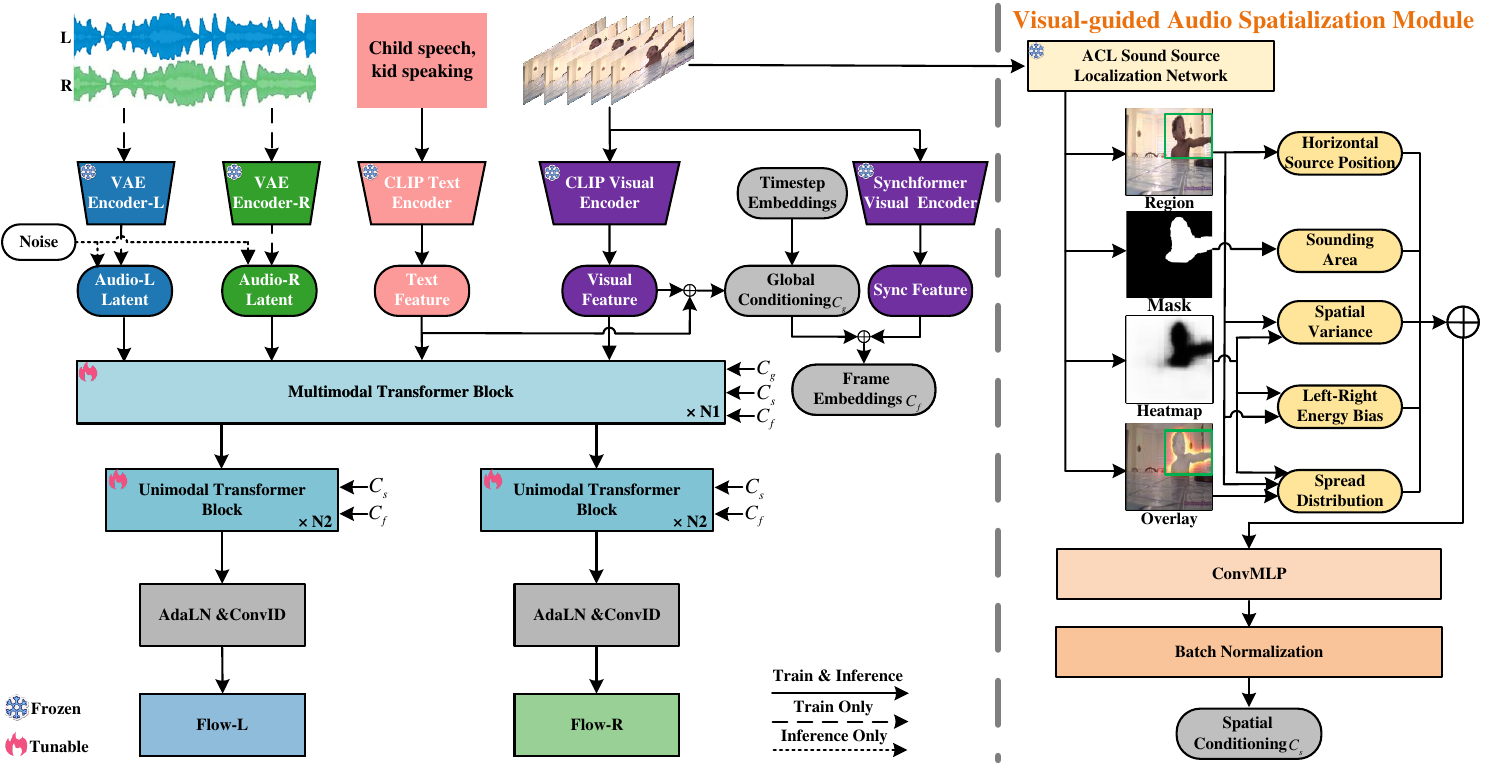}
\caption{
Overall architecture of SpatialV2A. 
The framework employs a dual-branch design for spatial audio generation (Sec. \ref{Dual_gen_frame}) and integrates a visual-guided audio spatialization module (Sec. \ref{spatial_module}) that injects video-derived spatial cues to enhance spatial perception.}
\label{SpatialV2A}
\end{figure*}

\section{Method}
\label{sec:method}

We propose \textbf{SpatialV2A}, a dual-branch spatial V2A framework built upon \textit{Conditional Flow Matching (CFM)} and enhanced with visual-guided spatial cues (as shown in Figure \ref{SpatialV2A}).
The framework first learns a \textit{multimodal joint representation} to align semantics across modalities and provide conditioning for binaural audio generation.
Building upon this shared representation, 
a \textit{unimodal dual-channel representation} decouples  the synthesis process into separate left and right branches to model channel-specific spatial variations.
A dedicated \textit{visual-guided audio spatialization} module further extracts spatial cues from the video to guide binaural channel differentiation.

\subsection{Preliminaries}
CFM is a continuous-time conditional generative framework that learns a condition-dependent velocity field to transport samples from a simple prior distribution to the target data distribution under given conditions \cite{tong2024improving}.
Given a target data sample $x_1$ and a noise sample $x_0\sim \mathcal{N}(0, I)$, an intermediate state is defined as $x_t=(1-t) x_0+t x_1, t \in[0,1]$.
The objective of CFM is to learn a conditional velocity field $v_\theta(x_t, t, \mathbf{C})$ parameterized by $\theta$, 
where $\mathbf{C}$ denotes the conditioning information (\textit{e.g.}, multimodal or unimodal features).
This is achieved by minimizing the loss:
\begin{equation}
\mathbb{E}_{x_0,x_1,t, \mathbf{C}}\left\|v_\theta\left(x_t,t,\mathbf{C}\right)-u\left(x_t \mid x_0, x_1\right)\right\|^2\text{,}
\end{equation}
where $u\left(x_t \mid x_0, x_1\right)$ is the ground-truth conditional velocity.
%


To generate binaural audio, 
we extend the standard CFM framework to explicitly model the left and right audio channels. 
Instead of learning a single velocity field, 
our framework learns two channel-specific velocity fields,
$v_\theta^{l}(x_t^{l}, t, \mathbf{C})$ and $v_\theta^{r}(x_t^{r}, t, \mathbf{C})$.
%
Given a binaural target audio sample $x_1 = (x_1^{l}, x_1^{r})$ from the constructed BinauralVGGSound dataset and the corresponding noise samples $x_0 = (x_0^{l}, x_0^{r})$, the intermediate states for each channel are constructed via linear interpolation as
$x_t^{a} = (1 - t)x_0^{a} + t x_1^{a}$, where $a \in \{l, r\}$.
Accordingly, the binaural conditional flow matching objective is:
\begin{equation}
\sum_{a \in\{l, r\}} \mathbb{E}_{x_0^{a},x_1^{a},t, \mathbf{C}}\left\|v_\theta^{a}\left(x_t^{a},t,\mathbf{C}\right)-u\left(x_t^{a} \mid x_0^{a}, x_1^{a}\right)\right\|^2 \text{.}
\end{equation}


\subsection{Dual-Branch Spatial Audio Generation}
\label{Dual_gen_frame}

\subsubsection{Multimodal Joint Representation}
\label{Multimodal_block}
To achieve cross-modal alignment and provide shared conditioning for binaural audio generation, we adopt multimodal joint transformer blocks following the design of MMAudio \cite{cheng2025mmaudio}.
We leverage textual $F_\text{text}$, visual $F_\text{vis}$, and synchronization $F_\text{sync}$ features for semantic grounding and temporal alignment, implemented via pretrained CLIP \cite{radford2021learning} and Synchformer \cite{iashin2024synchformer}.
The latent representations ($x_t^{l}$, $x_t^{r}$) are jointly updated by conditioning on six components $\{F_{\text{text}},\, F_{\text{vis}},\, F_{\text{sync}},\, C_s,\, C_f,\, C_g\}$.
The global conditioning $C_g$, shared across all transformer blocks, is $C_g = F_{\text{text}} \oplus F_{\text{vis}} \oplus e(t)$,
where $e(t)$ denotes the Fourier encoding of the flow timestep.
The frame-level conditioning is $C_f = F_{\text{sync}} \oplus  C_g$.
The spatial conditioning $C_s$ is detailed in Sec. \ref{spatial_module}.

\subsubsection{Unimodal Dual-Channel Representation}
\label{Single_block}
To model channel-specific spatial characteristics, we introduce unimodal dual-channel blocks that decouple generation into left and right branches after the multimodal stage. 
%
%
Each branch is implemented by a unimodal transformer \cite{labs2025flux1kontextflowmatching} to support independent spatial synthesis while maintaining architectural stability.
At each timestep, each branch's the latent representation is updated by conditioning on the $C_f$ and $C_s$.

\begin{table*}[htbp]
  \centering
  \caption{Quantitative and qualitative evaluation across distribution matching, temporal alignment, and semantic alignment. \textbf{Bold} indicates the best performance; \underline{underlined} denotes the second best.}
  \resizebox{\textwidth}{!}{
  \renewcommand{\arraystretch}{1.2} 
    \begin{tabular}{lcccccccc}
    \toprule[1.5pt]
   Method  & \multicolumn{1}{c}{$\text{KL}_\text{PaSST}$ ↓} & \multicolumn{1}{c}{$\text{KL}_\text{PANNs}$ ↓} & \multicolumn{1}{c}{$\text{FAD}_\text{PaSST}$ ↓} & \multicolumn{1}{c}{$\text{FAD}_\text{PaNNs} $ ↓} & \multicolumn{1}{c}{IB ↑} & \multicolumn{1}{c}{DeSync ↓} & \multicolumn{1}{c}{MOS-Sem ↑} & \multicolumn{1}{c}{MOS-T ↑} \\
    \hline
    ReWaS & 2.47  & 2.51  & 693.89  & 23.46  & 0.18  & 1.19  &  2.31 ± 0.62  &  2.21 ± 0.44\\
    Frieren & 2.62  & 2.60  & 490.00  & 16.96  & 0.22  & 0.85  &   3.40 ± 0.50    & 2.78 ± 0.58 \\
    Seeing\&Hearing & 2.66  & 2.78  & 755.96  & 28.79  & \textbf{0.36} & 1.20  &   2.60 ± 0.60    & 2.53 ± 0.93 \\
    MMAudio & \underline{1.57}  & \underline{1.59}  & \textbf{303.92} & \textbf{8.05} & \underline{0.29}  & \underline{0.51}  &  \textbf{4.11 ± 0.31}   & \underline{4.05 ± 0.60} \\
    SEE-2-SOUND & 3.36  & 4.39  & 3807.54  & 128.03  & 0.07  & 1.27  &     1.60 ± 0.72  &  1.51 ± 0.66\\
    \textbf{SpatialV2A(ours)} & \textbf{1.56} & \textbf{1.55} & \underline{381.90}  & \underline{12.53}  & 0.28  & \textbf{0.49} &   \underline{4.08 ± 0.36}    &  \textbf{4.14 ± 0.54}\\
    \bottomrule[1.5pt]
    \end{tabular}%
    }
  \label{Table_dis_tem_sem}%
\end{table*}%

\subsection{Visual-guided Audio Spatialization}
\label{spatial_module}
This module is designed to provide spatial conditioning $C_g$ for SpatialV2A. 
Using the pretrained audio-visual localization method ACL \cite{park2025hearing}, we extract the sounding region, mask, and heatmap from which five key spatial features are computed.

\textbf{1. Horizontal sound position} $S_h$ represents the relative horizontal placement of the sound source, and is defined as ${S}_h=\frac{cx}{W}$, where $cx$ denotes the horizontal centroid of the heatmap, and $W$ is the width of the heatmap.

\textbf{2. Sounding area fraction} $S_\text{area}$ quantifies the proportion of the sounding region within the heatmap and is defined as $S_\text{area} = \frac{\sum\nolimits_{x,y} \text{mask}_{x,y}}{H \times W}$,
where $x$ and $y$ denote pixel coordinates, and $\text{mask}$ is a binary array where pixels above a predefined threshold are marked as 1 and 0 otherwise, and $H$ denotes the height of the heatmap.

\textbf{3. Heatmap variance} $S_{var}$ measures the spatial variance within the heatmap, reflecting the dispersion of the sound source in the spatial domain. 
It is computed as the sum of the variance in the horizontal $\operatorname{var}_{x}=\frac{\sum\nolimits_{x}{(M(x)\cdot {{(x-cx)}^{2}})}}{\sum\nolimits_{x}{M(x)}}$ and vertical ${{\operatorname{var}}_{y}}=\frac{\sum\nolimits_{y}{(M(y)\cdot {{(y-cy)}^{2}})}}{\sum\nolimits_{y}{M(y)}}$ directions.
The overall spatial variance is then given by
${S}_{\operatorname{var}}=\operatorname{var}_{x}+{\operatorname{var}}_{y}$.
Where $cy$ is the vertical centroid of the heatmap, $M(\cdot)$ represents the localization heatmap predicted by ACL.

\textbf{4. Left-right energy bias} $S_\text{LR}$ captures the energy imbalance between the left and right halves of the visual frame:
\begin{equation}
{{S}_{\text{LR}}}=\frac{\sum\limits_{x=W/2+1}^{W}{\sum\limits_{y=1}^{H}{M(x,y)}}-\sum\limits_{x=1}^{W/2}{\sum\limits_{y=1}^{H}{M(x,y)}}}{\sum\limits_{x=1}^{W}{\sum\limits_{y=1}^{H}{M(x,y)}}}\text{.}
\end{equation}


\textbf{5. Spread distribution shape} $S_\text{shape}$ characterizes the anisotropy of the sound source by measuring its spread along the horizontal and vertical axes.
It is defined as $ {S}_\text{shape}=\frac{\operatorname{var}_{x}}{{\operatorname{var}}_{y}}$.


These features are concatenated into $S_{\text{sound}}$,
projected into a latent space via a Convolutional Multi-Layer Perceptron (ConvMLP) with upsampling, and normalized to form the spatial conditioning:
%
\begin{equation}
\begin{gathered}
S_{\text{sound}} = S_h \oplus S_{\text{area}} \oplus S_{\text{var}} \oplus S_{\text{LR}} \oplus S_{\text{shape}}\text{,}\\
C_s = \text{LayerNorm}\big( \text{ConvMLP}( \text{UpSample}( S_{\text{sound}} ) ) \big)\text{.}
\end{gathered}
\end{equation}

The high-dimensional spatial features are incorporated into the flow-matching process of a multimodal transformer, ensuring consistent feature scaling and training stability. 
This design jointly encodes sound source location, extent, dispersion, energy bias, and directionality, providing essential spatial conditions for binaural audio generation and localization.

\section{Experiments}
\label{sec:experiment}

\subsection{Experimental Details}

\textbf{Training and Inference.}
We train our SpatialV2A model on the BinauralVGGSound dataset.
During training, audio waveforms are first encoded into latent representations using a pretrained VAE \cite{kingma2013auto}.
The model is then optimized with the binaural CFM objective (defined in the Preliminaries) to learn the conditional velocity field between noise and target audio latents.
During inference, the left and right audio channels are both initialized from Gaussian noise and progressively transformed into binaural audio latents.
These predicted latents are decoded into mel-spectrograms via the pretrained VAE and subsequently converted into waveform audio using a BigVGAN vocoder \cite{lee2022bigvgan}.

\textbf{Implementation Details.}
The model is optimized using AdamW \cite{kingma2015adam} with a base learning rate of $1\times10^{-4}$, a linear warm-up over the first 1K steps, and a total of 300K training iterations. 
Our architecture employs $N_1 = 6$ multimodal joint transformer blocks and $N_2 = 12$ unimodal dual-channel blocks.
Audio is generated at a sampling rate of 16\,kHz and represented as 20-dimensional latent codes at 31.25\,fps \cite{jeong2025read,wang2024frieren}.
All experiments are conducted on three NVIDIA RTX 6000 GPUs with 96 GB of memory.

\textbf{Baselines.}
We selected several SOTA methods in the V2A as our baselines, 
including monaural generation models MMAudio \cite{jeong2025read}, ReWaS \cite{jeong2025read}, Frieren \cite{wang2024frieren}, and Seeing-and-Hearing \cite{xing2024seeing}, as well as the spatial generation model SEE-2-SOUND \cite{dagli2025see}.
Notably, all baseline models were originally trained on the VGGSound dataset.
For a fair evaluation, we evaluate all methods on the same subset of 3,000 videos randomly sampled from the original VGGSound test split, which shares identical video IDs with BinauralVGGSound.

\begin{table*}[htbp]
  \centering
  \caption{Quantitative and qualitative evaluation of audio quality and spatial alignment.}
  \resizebox{\textwidth}{!}{
  \renewcommand{\arraystretch}{1.1} 
    \begin{tabular}{lcccccccccccc}
        \toprule[1.5pt]
    Method & $\text{IS}_\text{PaSST}$ ↑ & $\text{IS}_\text{PaNNs}$ ↑ & CE ↑  & CU ↑  & PQ ↑  & IACC ↓ & ILD ↑ & ITD ↑ & ISD ↑ & IPD ↑ & MOS-AQ ↑& MOS-S ↑\\
    \hline
    ReWaS  & 7.86  & 8.76  & 3.54  & 4.82  & 5.34  & -     & -     & -     & -     & -     & 2.29 ± 0.62 & 1.26 ± 0.44 \\
    Frieren  & 9.10  & 8.69  & 3.65  & \underline{5.26}  & \underline{5.85}  & -     & -     & -     & -     & -     & 3.15 ± 0.52 & 1.29 ± 0.46 \\
    Seeing\&Hearing  & 5.39  & 5.19  & 3.06  & 4.49  & 5.25  & -     & -     & -     & -     & -     & 2.42 ± 0.63 & 1.50 ± 0.54 \\
    MMAudio & \underline{11.86}  & \underline{12.47}  & \underline{3.88}  & 5.14  & 5.71  & -     & -     & -     & -     & -     & \underline{3.87 ± 0.78} & 1.78 ± 0.61 \\
    SEE-2-SOUND  & 1.52  & 1.52  & 3.03  & 4.83  & 5.39  & -     & -     & -     & -     & -     & 1.74 ± 0.78 & \underline{2.49 ± 0.73} \\
    \textbf{SpatialV2A(ours)} & \textbf{14.36} & \textbf{14.91} & \textbf{4.09} & \textbf{5.77} & \textbf{6.07} & \textbf{0.51}  & \textbf{5.69}  & \textbf{1.79}  & \textbf{0.49}  & \textbf{1.12}  & \textbf{4.21 ± 0.43} & \textbf{3.90 ± 0.53} \\
    \bottomrule[1.5pt]
    \end{tabular}%
    }
  \label{table_SQ_spatial}%
\end{table*}%

\begin{figure}[H]
\centering
\includegraphics[width=0.45\textwidth]{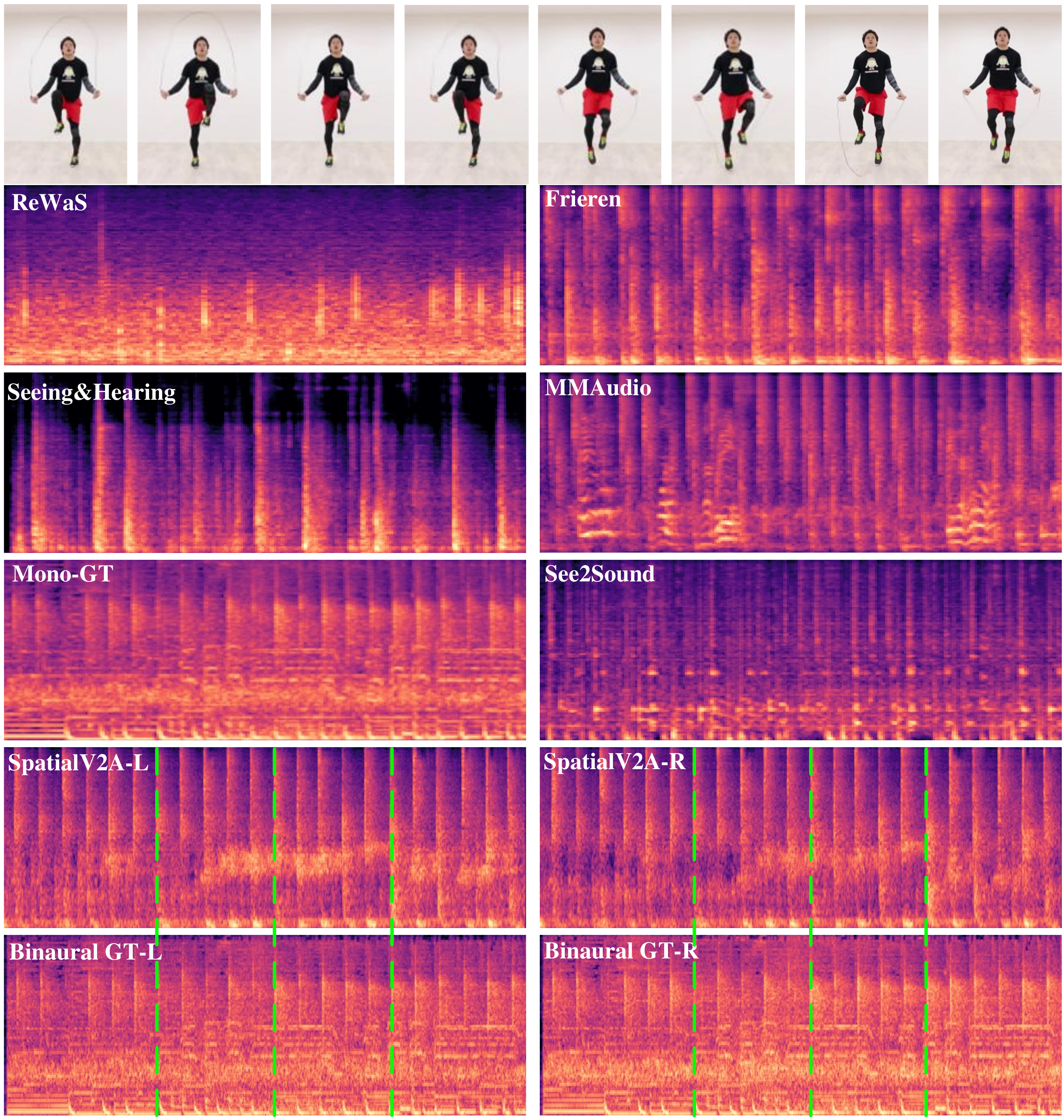}
\caption{Mel-spectrogram comparison between our method, baselines, original mono GT, and our binaural GT.
Our approach better preserves salient audio effects while reducing background noise.}
\label{mel_comparison}
\end{figure}

\subsection{Evaluation Metrics}

We evaluate the generated audio across five complementary dimensions. 

\textbf{Distribution Matching} quantifies the similarity between the feature distributions of GT and generated audio, reflecting overall fidelity.
We compute Fréchet Audio Distance (FAD) and Kullback–Leibler (KL) divergence using PaSST \cite{koutini2021efficient} and PANNs \cite{kong2020panns} as feature extractors.

\textbf {Semantic and Temporal Alignments} capture the consistency between audio and visual modalities in terms of semantics and temporal synchronization.
We employ ImageBind (IB) \cite{viertola2025temporally} and DeSync \cite{iashin2024synchformer} to estimate semantic and temporal consistency, respectively.

\textbf{Audio Quality} evaluates the perceptual clarity and naturalness of the generated sound using PQ, CE, and CU, together with the Inception Score (IS) \cite{salimans2016improved} computed using PaSST and PANNs. 

\textbf {Spatial Alignment} assesses spatial characteristics and localization accuracy in binaural generation. 
We measure ILD, ITD, ISD, IPD, and IACC \cite{hernandez2024interaural}.

\textbf{Subjective Evaluation.} We conducted a MOS study (1 = poor, 5 = excellent) with twenty invited participants who rated audio samples on high-quality headphones across four dimensions: Spatial Impression (MOS-S), Temporal Alignment (MOS-T), Semantic Alignment (MOS-Sem), and Audio Quality (MOS-AQ).

\subsection{Main Results}

Table \ref{Table_dis_tem_sem} shows the performance of baselines and the proposed method in terms of distribution matching, as well as temporal and semantic alignment.
We observe that SpatialV2A achieves the lowest KL divergence and DeSync scores, and ranks first or second on all distribution-related metrics, achieving the highest MOS-T and the second-highest MOS-Sem.
It is worth noting that Seeing\&Hearing method achieves the highest IB score because it directly optimizes the ImageBind similarity during inference.

Beyond these aspects, Table~\ref{table_SQ_spatial} reports results on audio quality and spatial perception. 
SpatialV2A achieves the best performance across both objective and subjective metrics.
Specifically, it attains the highest scores on audio quality metrics while effectively introducing spatial cues, as evidenced by performance on spatial metrics.
These spatial metrics are not applicable for all baseline methods that generate mono or non-spatial audio.
Moreover, SpatialV2A reaches the highest MOS-AQ (4.21) and MOS-S (3.90), significantly outperforming prior works in perceived sound quality and spatial realism. 
Notably, while monaural audio may convey a sense of immersion or depth through loudness variation or temporal dynamics, it lacks explicit interaural cues required for true spatialization, such as interaural level and time differences.

Figure \ref{mel_comparison} presents a mel-spectrogram comparison between SpatialV2A and baseline methods. 
As observed, our approach successfully reconstructs the audio corresponding to the real-world jump-rope scene, exhibiting a more concentrated energy distribution and stable time–frequency structure, which typically indicates higher perceptual clarity and reduced noise interference. 

Overall, these results confirm that SpatialV2A significantly enhances audio quality and spatial perception while maintaining strong temporal and semantic alignment with visual content.

\subsection{Ablation Study}

\begin{table}[t]
  \centering
  \caption{Ablation study on the proposed visual-guided Audio Spatialization module. We compare our full model (\textbf{SpatialV2A}) with a variant without spatialization (\textbf{Ours w/o Spat.}). }
  \resizebox{0.49\textwidth}{!}{
  \renewcommand{\arraystretch}{1.2} 
    \begin{tabular}{lcccccrrr}
    \toprule[1.5pt]
  Method & $\text{FAD}_\text{PANNs}$ ↓ & IB ↑  & DeSync ↓ & $\text{IS}_\text{PANNs}$ ↑ & IACC ↓ & \multicolumn{1}{c}{ILD ↑}  \\
  \hline
   \textbf{Ours w/o Spat.} & \textbf{11.01} 
  & 0.26  & 0.56  & 13.41  & 0.53  & 5.55     \\
   \textbf{SpatialV2A} & 12.53 
 & \textbf{0.28} & \textbf{0.49} & \textbf{14.91} & \textbf{0.51}  & \textbf{5.69}     \\
   \bottomrule[1.5pt]
    \end{tabular}%
    }
  \label{tab:Ablation}%
\end{table}%

To validate the contribution of the proposed \textit{Visual-guided Audio Spatialization} module, we conduct an ablation study by comparing the full SpatialV2A model with a variant that disables spatial conditioning injection (Ours w/o Spat.).
As shown in Table \ref{tab:Ablation}, removing spatialization slightly improves FAD, indicating marginally improved global distribution matching. 
In contrast, the full SpatialV2A model achieves higher performance across all critical dimensions, including semantic alignment (IB), temporal synchronization (DeSync), perceptual quality (IS), and binaural spatial metrics (IACC, ILD).
These results demonstrate that visual-guided spatialization, despite introducing minor distributional variation, substantially improves temporal coherence, semantic fidelity, and spatial realism.

\section{Conclusion}
\label{sec:conclusion}
This work addresses the overlooked limitation of spatial perception in video-to-audio generation. 
We introduce BinauralVGGSound, the first large-scale video-binaural audio dataset, and SpatialV2A, a novel dual-branch framework for visual-guided spatial audio synthesis.
Experimental results demonstrate that our approach generates high-fidelity binaural audio with accurate semantic, temporal, and spatial alignment.
This significantly enhances the immersive quality and perceptual realism of the synthesized soundscape.
For future work, we plan toincorporate captioned datasets such as WavCaps \cite{mei2024wavcaps} and AudioCaps \cite{kim2019audiocaps} to strengthen semantic grounding, and to explore advanced architectures for long-term temporal modeling, enabling coherent binaural synthesis in extended video sequences.

\bibliographystyle{named}
\bibliography{ijcai26}

@inproceedings{huang2025spotlighting,
  title={Spotlighting Partially Visible Cinematic Language for Video-to-Audio Generation via Self-distillation},
  author={Huang, Feizhen and Wu, Yu and Lin, Yutian and Du, Bo},
  booktitle={Proceedings of the Thirty-Fourth International Joint Conference on Artificial Intelligence},
  pages={1170--1178},
  year={2025}
}

@inproceedings{pan2024innovative,
  title={Innovative directional encoding in speech processing: leveraging spherical harmonics injection for multi-channel speech enhancement},
  author={Pan, Jiahui and Shen, Pengjie and Zhang, Hui and Zhang, Xueliang},
  booktitle={Proceedings of the Thirty-Third International Joint Conference on Artificial Intelligence},
  pages={6451--6459},
  year={2024}
}

@inproceedings{huang2023discovering,
  title={Discovering Sounding Objects by Audio Queries for Audio-Visual Segmentation},
  author={Huang, Shaofei and Li, Han and Wang, Yuqing and Zhu, Hongji and Dai, Jiao and Han, Jizhong and Rong, Wenge and Liu, Si},
  booktitle={Proceedings of the Thirty-Second International Joint Conference on Artificial Intelligence},
  pages={875--883},
  year={2023}
}

@inproceedings{jeong2025read,
  title={Read, watch and scream! sound generation from text and video},
  author={Jeong, Yujin and Kim, Yunji and Chun, Sanghyuk and Lee, Jiyoung},
  booktitle={Proceedings of the AAAI Conference on Artificial Intelligence},
  volume={39},
  number={17},
  pages={17590--17598},
  year={2025}
}

@article{luo2023diff,
  title={Diff-foley: Synchronized video-to-audio synthesis with latent diffusion models},
  author={Luo, Simian and Yan, Chuanhao and Hu, Chenxu and Zhao, Hang},
  journal={Advances in Neural Information Processing Systems},
  volume={36},
  pages={48855--48876},
  year={2023}
}

@inproceedings{kingma2013auto,
  title={Auto-Encoding Variational Bayes},
  author={Diederik P Kingma and Max Welling},
  booktitle={International Conference on Learning Representations},
  year={2014},
  url={https://openreview.net/forum?id=33X9fd2-9FyZd}
}

@inproceedings{sheffer2023hear,
  title={I hear your true colors: Image guided audio generation},
  author={Sheffer, Roy and Adi, Yossi},
  booktitle={IEEE International Conference on Acoustics, Speech and Signal Processing},
  pages={1--5},
  year={2023}
}

@inproceedings{lee2022bigvgan,
  title = {BigVGAN: A Universal Neural Vocoder with Large-Scale Training},
  author = {Lee, Sang-gil and Ping, Wei and Ginsburg, Boris and Catanzaro, Bryan and Yoon, Sungroh},
  booktitle = {International Conference on Learning Representations},
  year = {2023}
}

@inproceedings{chen2020vggsound,
  title={Vggsound: A large-scale audio-visual dataset},
  author={Chen, Honglie and Xie, Weidi and Vedaldi, Andrea and Zisserman, Andrew},
  booktitle={IEEE International Conference on Acoustics, Speech and Signal Processing},
  pages={721--725},
  year={2020}
}

@article{garg2023visually,
  title={Visually-guided audio spatialization in video with geometry-aware multi-task learning},
  author={Garg, Rishabh and Gao, Ruohan and Grauman, Kristen},
  journal={International Journal of Computer Vision},
  volume={131},
  number={10},
  pages={2723--2737},
  year={2023}
}

@inproceedings{cheng2025mmaudio,
  title={MMAudio: Taming Multimodal Joint Training for High-Quality Video-to-Audio Synthesis},
  author={Cheng, Ho Kei and Ishii, Masato and Hayakawa, Akio and Shibuya, Takashi and Schwing, Alexander and Mitsufuji, Yuki},
  booktitle={Proceedings of the IEEE/CVF Conference on Computer Vision and Pattern Recognition},
  pages={28901--28911},
  year={2025}
}

@misc{labs2025flux1kontextflowmatching,
      title={FLUX.1 Kontext: Flow Matching for In-Context Image Generation and Editing in Latent Space},
      author={Black Forest Labs and Stephen Batifol and Andreas Blattmann and Frederic Boesel and Saksham Consul and Cyril Diagne and Tim Dockhorn and Jack English and Zion English and Patrick Esser and Sumith Kulal and Kyle Lacey and Yam Levi and Cheng Li and Dominik Lorenz and Jonas Müller and Dustin Podell and Robin Rombach and Harry Saini and Axel Sauer and Luke Smith},
      year={2025},
      eprint={2506.15742},
      archivePrefix={arXiv},
      primaryClass={cs.GR},
      url={https://arxiv.org/abs/2506.15742},
}

@inproceedings{iashin2024synchformer,
  title={Synchformer: Efficient synchronization from sparse cues},
  author={Iashin, Vladimir and Xie, Weidi and Rahtu, Esa and Zisserman, Andrew},
  booktitle={IEEE International Conference on Acoustics, Speech and Signal Processing},
  pages={5325--5329},
  year={2024},
  organization={IEEE}
}

@inproceedings{radford2021learning,
  title={Learning transferable visual models from natural language supervision},
  author={Radford, Alec and Kim, Jong Wook and Hallacy, Chris and Ramesh, Aditya and Goh, Gabriel and Agarwal, Sandhini and Sastry, Girish and Askell, Amanda and Mishkin, Pamela and Clark, Jack and others},
  booktitle={International Conference on Machine Learning},
  pages={8748--8763},
  year={2021}
}

@article{wang2024frieren,
  title={Frieren: Efficient video-to-audio generation network with rectified flow matching},
  author={Wang, Yongqi and Guo, Wenxiang and Huang, Rongjie and Huang, Jiawei and Wang, Zehan and You, Fuming and Li, Ruiqi and Zhao, Zhou},
  journal={Advances in Neural Information Processing Systems},
  volume={37},
  pages={128118--128138},
  year={2024}
}

@inproceedings{dagli2025see,
  title={See-2-sound: Zero-shot spatial environment-to-spatial sound},
  author={Dagli, Rishit and Prakash, Shivesh and Wu, Robert and Khosravani, Houman},
  booktitle={Proceedings of the Special Interest Group on Computer Graphics and Interactive Techniques Conference},
  pages={1--2},
  year={2025}
}

@article{zhang2025visaudio,
  title={ViSAudio: End-to-End Video-Driven Binaural Spatial Audio Generation},
  author={Zhang, Mengchen and Chen, Qi and Wu, Tong and Liu, Zihan and Lin, Dahua},
  journal={arXiv preprint arXiv:2512.03036},
  year={2025}
}

@article{xumanjie2024video,
  title={Video-to-audio generation with hidden alignment},
  author={Xu, Manjie and Li, Chenxing and Tu, Xinyi and Ren, Yong and Chen, Rilin and Gu, Yu and Liang, Wei and Yu, Dong},
  journal={arXiv preprint arXiv:2407.07464},
  year={2024}
}

@inproceedings{xing2024seeing,
  title={Seeing and hearing: Open-domain visual-audio generation with diffusion latent aligners},
  author={Xing, Yazhou and He, Yingqing and Tian, Zeyue and Wang, Xintao and Chen, Qifeng},
  booktitle={Proceedings of the IEEE/CVF Conference on Computer Vision and Pattern Recognition},
  pages={7151--7161},
  year={2024}
}

@inproceedings{koutini2021efficient,
  author={Koutini, Khaled and Schl{\"u}ter, Jan and Eghbal-Zadeh, Hamid and Widmer, Gerhard},
  title= {Efficient Training of Audio Transformers with Patchout},
  booktitle={International Speech Communication Association Conference},
  pages = {2753--2757},
  year= {2022}
}

@article{kong2020panns,
  title={Panns: Large-scale pretrained audio neural networks for audio pattern recognition},
  author={Kong, Qiuqiang and Cao, Yin and Iqbal, Turab and Wang, Yuxuan and Wang, Wenwu and Plumbley, Mark D},
  journal={IEEE/ACM Transactions on Audio, Speech, and Language Processing},
  volume={28},
  pages={2880--2894},
  year={2020}
}

@inproceedings{sun2023learning,
  title={Learning audio-visual source localization via false negative aware contrastive learning},
  author={Sun, Weixuan and Zhang, Jiayi and Wang, Jianyuan and Liu, Zheyuan and Zhong, Yiran and Feng, Tianpeng and Guo, Yandong and Zhang, Yanhao and Barnes, Nick},
  booktitle={Proceedings of the IEEE/CVF Conference on Computer Vision and Pattern Recognition},
  pages={6420--6429},
  year={2023}
}

@article{park2025hearing,
  title={Hearing and Seeing Through CLIP: A Framework for Self-Supervised Sound Source Localization},
  author={Park, Sooyoung and Senocak, Arda and Chung, Joon Son},
  journal={International Journal of Computer Vision},
  year={2025}
}

@article{salimans2016improved,
  title={Improved techniques for training gans},
  author={Salimans, Tim and Goodfellow, Ian and Zaremba, Wojciech and Cheung, Vicki and Radford, Alec and Chen, Xi},
  journal={Advances in Neural Information Processing Systems},
  volume={29},
  year={2016}
}

@inproceedings{kingma2015adam,
  title={Adam: A Method for Stochastic Optimization},
  author={Kingma, Diederik P. and Ba, Jimmy},
  booktitle={International Conference on Learning Representations},
  year={2015}
}

@article{mei2024wavcaps,
  title={Wavcaps: A chatgpt-assisted weakly-labelled audio captioning dataset for audio-language multimodal research},
  author={Mei, Xinhao and Meng, Chutong and Liu, Haohe and Kong, Qiuqiang and Ko, Tom and Zhao, Chengqi and Plumbley, Mark D and Zou, Yuexian and Wang, Wenwu},
  journal={IEEE/ACM Transactions on Audio, Speech, and Language Processing},
  volume={32},
  pages={3339--3354},
  year={2024}
}

@inproceedings{kim2019audiocaps,
  title={Audiocaps: Generating captions for audios in the wild},
  author={Kim, Chris Dongjoo and Kim, Byeongchang and Lee, Hyunmin and Kim, Gunhee},
  booktitle={Proceedings of the 2019 Conference of the North American Chapter of the Association for Computational Linguistics: Human Language Technologies},
  pages={119--132},
  year={2019}
}

@article{tjandra2025meta,
  title={Meta audiobox aesthetics: Unified automatic quality assessment for speech, music, and sound},
  author={Tjandra, Andros and Wu, Yi-Chiao and Guo, Baishan and Hoffman, John and Ellis, Brian and Vyas, Apoorv and Shi, Bowen and Chen, Sanyuan and Le, Matt and Zacharov, Nick and others},
  journal={arXiv preprint arXiv:2502.05139},
  year={2025}
}

@article{tong2024improving,
  title={Improving and generalizing flow-based generative models with minibatch optimal transport},
  author={Tong, Alexander and Fatras, Kilian and Malkin, Nikolay and Huguet, Guillaume and Zhang, Yanlei and Rector-Brooks, Jarrid and Wolf, Guy and Bengio, Yoshua},
  journal={Transactions on Machine Learning Research},
  issn={2835-8856},
  year={2024}
}

@inproceedings{hernandez2024interaural,
  title={Interaural time difference loss for binaural target sound extraction},
  author={Hernandez-Olivan, Carlos and Delcroix, Marc and Ochiai, Tsubasa and Tawara, Naohiro and Nakatani, Tomohiro and Araki, Shoko},
  booktitle={International Workshop on Acoustic Signal Enhancement},
  pages={210--214},
  year={2024}
}

@inproceedings{viertola2025temporally,
  title={Temporally aligned audio for video with autoregression},
  author={Viertola, Ilpo and Iashin, Vladimir and Rahtu, Esa},
  booktitle={IEEE International Conference on Acoustics, Speech and Signal Processing},
  pages={1--5},
  year={2025}
}

@inproceedings{zhou2020sep,
  title={Sep-stereo: Visually guided stereophonic audio generation by associating source separation},
  author={Zhou, Hang and Xu, Xudong and Lin, Dahua and Wang, Xiaogang and Liu, Ziwei},
  booktitle={Proceedings of the European Conference on Computer Vision},
  pages={52--69}
}

@inproceedings{gao20192,
  title={2.5 d visual sound},
  author={Gao, Ruohan and Grauman, Kristen},
  booktitle={Proceedings of the IEEE/CVF Conference on Computer Vision and Pattern Recognition},
  pages={324--333},
  year={2019}
}

@inproceedings{he2016deep,
  title={Deep residual learning for image recognition},
  author={He, Kaiming and Zhang, Xiangyu and Ren, Shaoqing and Sun, Jian},
  booktitle={Proceedings of the IEEE/CVF Conference on Computer Vision and Pattern Recognition},
  pages={770--778},
  year={2016}
}

@inproceedings{ronneberger2015u,
  title={U-net: Convolutional networks for biomedical image segmentation},
  author={Ronneberger, Olaf and Fischer, Philipp and Brox, Thomas},
  booktitle={International Conference on Medical image computing and computer-assisted intervention},
  pages={234--241},
  year={2015},
  organization={Springer}
}

@article{zhang2011high,
  title={On high-resolution head-related transfer function measurements: An efficient sampling scheme},
  author={Zhang, Wen and Zhang, Mengqiu and Kennedy, Rodney A and Abhayapala, Thushara D},
  journal={IEEE Transactions on Audio, Speech, and Language Processing},
  volume={20},
  number={2},
  pages={575--584},
  year={2011}
}

@inproceedings{xu2021visually,
  title={Visually informed binaural audio generation without binaural audios},
  author={Xu, Xudong and Zhou, Hang and Liu, Ziwei and Dai, Bo and Wang, Xiaogang and Lin, Dahua},
  booktitle={Proceedings of the IEEE/CVF Conference on Computer Vision and Pattern Recognition},
  pages={15485--15494},
  year={2021}
}

@inproceedings{CCStereo2025,
author = {Chen, Yuanhong and Shimada, Kazuki and Simon, Christian and Ikemiya, Yukara and Shibuya, Takashi and Mitsufuji, Yuki},
title = {CCStereo: Audio-Visual Contextual and Contrastive Learning for Binaural Audio Generation},
year = {2025},
booktitle = {Proceedings of the ACM International Conference on Multimedia},
pages = {7510–7518}
}

@InProceedings{iashin2021taming,
  title={Taming Visually Guided Sound Generation},
  author={Iashin, Vladimir and Rahtu, Esa},
  booktitle={British Machine Vision Conference},
  year={2021},
}

\end{document}